\documentclass{article}

\ifdefined\anonymous
  \usepackage[dblblindworkshop]{neurips_2026_vericode}
\else
  \usepackage[final]{neurips_2026_vericode}
\fi

\usepackage{float}

\usepackage[utf8]{inputenc} % allow utf-8 input
\usepackage[T1]{fontenc}    % use 8-bit T1 fonts
\usepackage{hyperref}       % hyperlinks
\usepackage{url}            % simple URL typesetting
\usepackage{multirow}            % simple URL typesetting

\usepackage{booktabs}       % professional-quality tables
\usepackage{amsfonts}       % blackboard math symbols
\usepackage{nicefrac}       % compact symbols for 1/2, etc.
\usepackage{microtype}      % microtypography
\usepackage{xcolor}         % colors
\usepackage{array}
\usepackage{placeins}
\usepackage{upquote}
\usepackage{amssymb}        % \checkmark in Figure 1
\usepackage{graphicx}
\usepackage{tikz}
\usetikzlibrary{positioning,arrows.meta,calc}
\definecolor{ccblue}{RGB}{37,99,235}    % Figure 1: natural-language requirement
\definecolor{ccamber}{RGB}{217,119,6}   % Figure 1: formal artifact
\definecolor{ccgreen}{RGB}{5,150,105}   % Figure 1: informalization
\definecolor{ccred}{RGB}{220,38,38}     % Figure 1: verdict, blocked edge
\definecolor{ccgray}{RGB}{107,114,128}  % Figure 1: reference edge

\newcommand{\stagehead}[2]{%
  \textsc{\footnotesize #1}\hfill{\footnotesize\itshape #2}\par\vspace{2pt}}
\newenvironment{trace}[1]{%
  \small\setlength{\tabcolsep}{0pt}%
  \begin{tabular}{@{}>{\raggedright\arraybackslash}p{\textwidth}@{}}
  \toprule
  #1\\
  \midrule
}{\bottomrule\end{tabular}}

\title{claimcheck: Measuring and Improving Specification Faithfulness}

\definecolor{darkorange}{RGB}{204,102,0}
\newcommand{\sh}[1]{\textcolor{red}{\textbf{SH:} #1}}
\newcommand{\hb}[1]{\textcolor{blue}{\textbf{HB:} #1}}
\newcommand{\oa}[1]{\textcolor{green!50!black}{\textbf{OA:} #1}}
\newcommand{\ma}[1]{\textcolor{purple}{\textbf{MA:} #1}}
\newcommand{\na}[1]{\textcolor{darkorange}{\textbf{NA:} #1}}
\ifdefined\anonymous  % margin comments do not appear in the submission build
  \renewcommand{\sh}[1]{}\renewcommand{\hb}[1]{}\renewcommand{\oa}[1]{}
  \renewcommand{\ma}[1]{}\renewcommand{\na}[1]{}
\fi

\author{%
  Martin Amster\thanks{Equal contribution, listed in alphabetical order; Apart Secure Program Synthesis fellows.}\\
  Universidad de Buenos Aires\\
  Buenos Aires, Argentina\\
  \texttt{mamster@dc.uba.ar} \\
  \And
  One An\footnotemark[1]\\
  University of Pennsylvania\\
  Philadelphia, PA 19104, USA\\
  \texttt{onean@engineering.upenn.edu}\\
  \And
  Holly Baker\footnotemark[1]\\
  PolicingLab\\
  Dublin, Ireland\\
 \texttt{baker@policinglab.org}\\
      \And
  Fernanda Graciolli\\
  Midspiral, Inc. \\
  Cambridge, MA 02139, USA \\
  \texttt{fernanda@midspiral.com} \\
    \AND
  Simon Henniger\thanks{Equal advising.} \\
  Harvard University \\
  Cambridge, MA 02138, USA \\
  \texttt{shenniger@fas.harvard.edu} \\
      \And
  Nada Amin\footnotemark[2] \\
  Harvard University \& Midspiral, Inc.\\
  Cambridge, MA 02138 \& MA 02139, USA \\
  \texttt{namin@seas.harvard.edu} \\
}

\begin{document}

\maketitle

\begin{abstract}
Formal verification proves that a program satisfies its specification, but says nothing about whether the specification captures what the author meant. As LLMs increasingly write both code and specifications, a gap between the author's natural-language requirement and the specification becomes a live risk. A
machine-checked proof of a property narrower than the one requested is worthless but passes every check. Closing this gap is solving the spec formalization problem: given a natural-language requirement, produce a formal artifact whose meaning matches it.

To measure the gap, we curate an end-to-end benchmark for spec
formalization spanning four formal languages: Dafny contracts and Lean~4, Isabelle and Rocq theorem statements. Each of its 3{,}234 rows pairs a natural-language requirement with a formal artifact and a label established by execution, by proof, or by the benchmark it is drawn from. A third of the rows are meaning-preserving rewrites of the artifact that keep the faithful label, so a judge cannot score by detecting that a row was edited.

We introduce \texttt{claimcheck}, a tool that detects incorrect formalizations: the formal artifact is translated back into plain English by a model that never sees the requirement, and a separate model, the judge, decides whether the informalization and the requirement express the same guarantee. On all four languages it gains $8.2$ to $9.8$ points of macro recall over a single call shown the requirement and the artifact, and $4.2$ to $6.0$ over that call handed the same list of meaning-preserving rewrites. Showing the judge the artifact as well costs $1.4$ to $4.5$ points. Across four judges that also see the artifact, the gain shrinks as the judge gets stronger, to within a point of the informed single call for the strongest.
\end{abstract}

\section{Introduction}

As models get better at writing code, the programmer's job shifts toward saying \emph{what} a program should do rather than \emph{how} it should do it. Spec-driven development already works this way: the developer states the requirement in English, and the implementation is generated from it. The hard part moves from writing the code to knowing whether it does what was asked.
For most of what a model produces there is no automatic way to tell, and checking falls to a human or to another model \citep{zheng2023llmjudge}. Formal languages are the exception. Here the requirement is first written down as a specification: a precise statement of what the program must guarantee, in a form a machine can check. A verifier then proves that the code meets that specification on every
possible input. But the verifier only ever sees the specification. It never sees the natural-language requirement that specification was written from, so it cannot tell whether the specification asked for the right thing.

Suppose the requirement is to sort a list. The specification says the output must be in increasing order, and the code is proved to satisfy it. Now take a program that ignores its input and returns an empty list. Nothing in an empty list is out of order, so that program meets the specification too, and the same proof would go through. The specification never said the output has to contain the same items as the input. This is the part no verifier covers. In our Dafny \citep{leino2010dafny} slice every row we labelled defective still passes verification, so a verified artifact can carry an assurance it has not earned. We call the task of getting it right \emph{spec formalization}: given a natural-language requirement, produce a formal artifact whose meaning matches it.

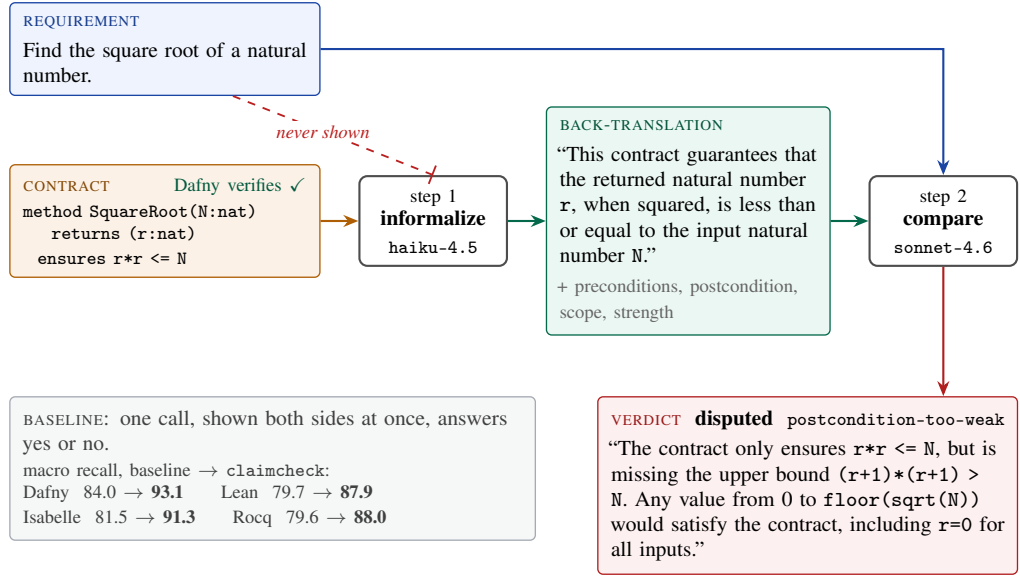
\begin{figure}[t]
\centering
% Figure 1: one claimcheck run on the integer_square_root C2 row, drawn as a
% pipeline. Requires: tikz with libraries arrows.meta, positioning, calc;
% xcolor. Colours: blue = informal English, amber = formal artifact,
% green = back-translation, red = verdict / blocked edge, gray = baseline note.
% The judge never sees the contract: the only path from the contract to the
% verdict runs through the informalization.
\providecommand{\fhead}[1]{{\scriptsize\scshape #1}}
\begin{tikzpicture}[
  font=\footnotesize,
  every node/.style={align=left},
  box/.style={draw, rounded corners=2pt, line width=0.5pt, inner sep=5pt, text width=#1},
  req/.style={box=#1, draw=ccblue!70!black, fill=ccblue!8},
  art/.style={box=#1, draw=ccamber!80!black, fill=ccamber!10},
  bt/.style={box=#1, draw=ccgreen!70!black, fill=ccgreen!8},
  ver/.style={box=#1, draw=ccred!80!black, fill=ccred!7},
  note/.style={box=#1, draw=ccgray!60, fill=ccgray!6, text=black!75},
  stage/.style={draw=black!70, rounded corners=3pt, line width=0.8pt, fill=white,
                inner sep=5pt, text width=#1, align=center},
  flow/.style={-{Stealth[length=2mm,width=1.6mm]}, line width=0.8pt},
  blocked/.style={-{Bar[width=2.6mm]}, dashed, ccred!85!black, line width=0.7pt},
  edgelab/.style={font=\scriptsize\itshape, fill=white, inner sep=1.5pt},
]

% ---- inputs ---------------------------------------------------------------
\node[art=3.75cm] (contract) at (0,0) {%
  \fhead{\textcolor{ccamber!60!black}{contract}}\hfill
  {\scriptsize\textcolor{ccgreen!60!black}{Dafny verifies \checkmark}}\\[2pt]
  {\ttfamily\scriptsize method SquareRoot(N:nat)\\
   \ \ \ \ returns (r:nat)\\
   \ \ ensures r*r <= N}};

\node[req=3.75cm, above=9mm of contract] (req) {%
  \fhead{\textcolor{ccblue!60!black}{requirement}}\\[2pt]
  Find the square root of a natural number.};

% ---- step 1 ---------------------------------------------------------------
\node[stage=1.6cm, right=5mm of contract] (inf) {%
  {\scriptsize step 1}\\[-1pt]
  \textbf{informalize}\\[1pt]
  {\scriptsize\ttfamily haiku-4.5}};

\node[bt=3.4cm, right=5mm of inf] (bt) {%
  \fhead{\textcolor{ccgreen!55!black}{back-translation}}\\[2pt]
  ``This contract guarantees that the returned natural number \texttt{r}, when
  squared, is less than or equal to the input natural number \texttt{N}.''\\[2pt]
  {\scriptsize\textcolor{black!60}{+ preconditions, postcondition, scope, strength}}};

% ---- step 2 ---------------------------------------------------------------
\node[stage=1.6cm, right=5mm of bt] (cmp) {%
  {\scriptsize step 2}\\[-1pt]
  \textbf{compare}\\[1pt]
  {\scriptsize\ttfamily sonnet-4.6}};

\coordinate (refy) at ([yshift=-4.5mm]bt.south); % below the tallest box in the row
\node[ver=5.2cm, anchor=north east] (verdict) at ([yshift=-3.5mm]refy -| cmp.east) {%
  \fhead{\textcolor{ccred!70!black}{verdict}}\hfill
  {\textbf{disputed}\ \ {\ttfamily\scriptsize postcondition-too-weak}}\\[2pt]
  ``The contract only ensures \texttt{r*r <= N}, but is missing the upper bound
  \texttt{(r+1)*(r+1) > N}. Any value from 0 to \texttt{floor(sqrt(N))} would
  satisfy the contract, including \texttt{r=0} for all inputs.''};

% ---- baseline contrast ----------------------------------------------------
\node[note=6.6cm, anchor=north west] (base) at ([yshift=-3.5mm]refy -| contract.west) {%
  \fhead{baseline}: one call, shown both sides at once, answers yes or no.\\[2pt]
  {\scriptsize macro recall, baseline $\rightarrow$ \texttt{claimcheck}:\\
   Dafny\ \ 84.0 $\rightarrow$ \textbf{93.1}\qquad Lean\ \ 79.7 $\rightarrow$ \textbf{87.9}\\
   Isabelle\ \ 81.5 $\rightarrow$ \textbf{91.3}\qquad Rocq\ \ 79.6 $\rightarrow$ \textbf{88.0}}};

% ---- edges ----------------------------------------------------------------
\draw[flow, ccamber!80!black] (contract) -- (inf);
\draw[flow, ccgreen!70!black] (inf) -- (bt);
\draw[flow, ccgreen!70!black] (bt) -- (cmp);

% requirement goes straight to the judge, never to the informalizer
\draw[flow, ccblue!70!black] (req.east) -| (cmp.north);
\draw[blocked] ([xshift=9mm]req.south) -- (inf.north)
  node[edgelab, text=ccred!85!black, pos=0.45] {never shown};

% verdict
\draw[flow, ccred!80!black] (cmp.south) -- (cmp.south |- verdict.north);

\end{tikzpicture}
\caption{One \texttt{claimcheck} run, on a Dafny contract from CloverBench
whose postcondition was weakened (class \texttt{C2}). Like the sorting
example, the contract still verifies and is satisfied by returning zero.
Step~1 translates the contract into English without ever seeing the
requirement, so the informalization can only report what the contract says.
Step~2 compares the informalization with the requirement and names the
discrepancy; it never sees the contract. Colours follow the data:
blue is the natural-language requirement, amber the formal artifact, green its
informalization. A single-call baseline shown both sides at once scores lower
on all four languages (Table~\ref{tab:results-main}).}
\label{fig:overview}
\end{figure}

We introduce \texttt{claimcheck}, a tool that checks whether a formal artifact means what its natural-language requirement asks for. The direct way to check this is to show a model both the requirement and the specification and ask whether they agree \citep{cabral2025proofflow, zhang2026beyondcompilation, lu2024formalalign}. The trouble is that the requirement gives the answer away. A model that has just read \emph{sort a list} expects the specification to describe sorting, and is biased toward agreeing that it does. Telling the model not to do this is an instruction it can fail to follow, so \texttt{claimcheck} separates reading the specification from judging it. A first model, the informalizer, puts the specification into plain English --- its preconditions, its postcondition, its scope --- without ever seeing the requirement, so what it writes can only report what the specification says. We call this the informalization \citep{wu2022autoformalization, amrollahi2026roundtrip}.
 A second model, the judge, then compares the informalization against the requirement (Figure~\ref{fig:overview}). What it compares is two English texts, one saying what was asked for and the other what the artifact guarantees, and the one it reads was written before the requirement could shape it. We also report the variant that hands the judge the artifact alongside the informalization; it is the weaker of the two on all four of our languages, by $1.4$ to $4.5$ points of macro recall.

\subsection{Contributions}
Our paper's main contributions are:
\begin{enumerate}
       \item We curate a test suite for end-to-end formalization of
    natural-language requirements, spanning four formal languages: Dafny
    contracts and Lean~4, Isabelle and Rocq theorem statements.
    Its 3{,}234 rows pair a requirement with a formal artifact and a label established by execution, by proof, or by the benchmark it is drawn from, and are scored by a standardized grader (Section~\ref{sec:benchmark}).
 Benchmarks of this kind
    are built by mutation, so their labels record whether a row was edited
    rather than what it now means. A third of our rows are control mutations,
    which rewrite the formal artifact while preserving its meaning and stay
    labelled faithful, so detecting edits is no longer enough to score.

\item We measure baseline scores for frontier models. A single-prompt judge shown the requirement and the artifact reaches between $79.6$ and $84.0$ macro recall across the four languages, and errs in both directions: on Dafny it disputes $20\%$ of the faithful contracts while missing a quarter of the rows whose requirement carries the defect.
    Handing that same single call the list of meaning-preserving rewrites and the error vocabulary \texttt{claimcheck}'s judge is given raises it by $3$ to $5$ points on all four languages, so we report it as a second baseline throughout: it is what separates the pipeline from the information in its prompt (Section~\ref{sec:results}).

    \item We introduce \texttt{claimcheck}, the two-step tool of
    Figure~\ref{fig:overview}. Comparing the two English texts gains $4.2$ to $6.0$ points of macro recall over that informed baseline on all four, most of it on the faithful rows, originals and controls alike. Three ablations locate the gain: removing the back-translation costs $2.8$ to $6.0$ points, giving the informalizer the requirement it is otherwise denied costs $0.8$ to $3.0$, and holding rewrites out of the judge's list moves macro recall by at most a point, so that list is not acting as an answer key for the benchmark's controls.
    Varying the judge, with the artifact in its view, shows the gain is substitutive rather than additive: up to $27$ points for the weakest model we test and none for the strongest
    (Section~\ref{sec:results}).
\end{enumerate}

\section{Benchmark Design}
\label{sec:benchmark}

Each row is a natural-language requirement, a formal artifact meant
to express it, and the context needed to read that artifact, and its label says whether the artifact means what the requirement asks for: \texttt{confirmed} or \texttt{disputed}. The benchmark is a union of four slices: 362 Dafny rows derived from CloverBench \citep{sun2024clover}, and 1{,}296 Lean~4, 920 Isabelle and 656 Rocq rows derived from PutnamBench \citep{tsoukalas2024putnambench}.

\paragraph{Meaning-preserving controls.} A faithfulness benchmark needs pairs
whose two sides are known to disagree, and the usual way to get them is
mutation. Take a correct pair, damage one of its two sides, and label the result
\texttt{disputed}. Both sides get damaged here. Clover mutates the contract in
two of its three variant classes and the docstring in the third; the PutnamBench
slices mutate the theorem statement. The construction has a defect that is easy
to miss. Suppose a judge ignores both sides of every row and answers
\texttt{confirmed} whenever the artifact looks untouched. It gets the entire
confirmed half right, because the label tracks where a row came from rather than
what it means. A high score on such a benchmark can therefore reflect nothing but
good mutation detection. We add a third family to close that hole. A
\emph{control mutation} rewrites the formal artifact while preserving its
meaning, and it stays labelled \texttt{confirmed}. Every slice carries one, and
together they are 1{,}029 of the 1{,}854 confirmed rows, so a judge betting on
provenance now loses most of the confirmed half either way. Controls carry a
burden that semantic mutations do not: a rewrite has to be \emph{shown} to
preserve meaning, not merely intended to, and the two subsections below set out
how each slice discharges it.

\subsection{Dafny}
CloverBench ships 62 hand-written textbook programs, each a docstring, which
serves as the requirement, paired with a contract, together with a taxonomy of
incorrect variants. We use three of its four released classes: the 62 originals
are labelled \texttt{confirmed}, and \texttt{C1}, which mutates the docstring
(62 rows), \texttt{C2}, which mutates the contract (60), and \texttt{C6}, which
mutates the contract and the body together (60), are \texttt{disputed}. All 182
defective rows still verify in Dafny, which the benchmark states as a design
requirement: the domain has a formal oracle and it reports success on every
defective row, so an autoformalization has to be checked for faithfulness even
where verification is available.

We add a fourth family of 118 control mutations over 45 of the 62 programs,
which rewrite the contract while preserving what it guarantees and so keep the
label \texttt{confirmed}. A candidate is kept only if Dafny proves the rewritten
contract equivalent to the original under its preconditions, and only if it
cannot also prove that postcondition false, which would make the equivalence
vacuous. The slice is 362 rows, 180 of them \texttt{confirmed}.

The formal artifact is the contract alone: declaration, \texttt{requires} and
\texttt{ensures}, with the body stripped. The body is what a contract is checked
\emph{against} rather than part of what it claims, and a judge given it could
infer the intent from the implementation instead of reading the contract.
Extraction is exact because no Clover program declares a helper function,
predicate or datatype. Four rows are dropped where the mutated contract retains
only framing clauses such as \texttt{modifies s}, which guarantee nothing and
leave nothing to compare, so \texttt{C2} and \texttt{C6} hold 60 rows each
(Appendix~\ref{app:variants}).

\subsection{Lean, Isabelle and Rocq}
\label{sec:putnam}

PutnamBench \citep{tsoukalas2024putnambench} formalizes the same competition
problems in all three languages, and the natural-language statement lives in one
file keyed by problem name, independent of the language. A row in these slices
therefore pairs the \emph{same} requirement with a different formal artifact,
which is what makes the three comparable: a difference between them is a
difference in the language, not in the question. The artifact is the theorem
statement and the requirement is the problem statement that sits above it in the
source, so the expression that extracts the one removes the other. PutnamBench
writes a problem's answer into a separate constant, which we substitute back
into the context before splitting the theorem, since a statement about an opaque
constant reads as vacuous whatever it says.

Five rewrite rules preserve meaning and four change it
(Table~\ref{tab:putnam-rules}), and the two families are warranted differently.
For the controls, Lean discharges the obligation \texttt{(original)}
$\leftrightarrow$ \texttt{(mutated)} per row in PutnamBench's own project under a
fixed tactic list, and a candidate that fails is dropped. In Isabelle and Rocq the
same five rules are propositional schemas --- $\top \wedge P \leftrightarrow P$,
$P \vee \bot \leftrightarrow P$, $\neg\neg P \leftrightarrow P$, a universal
through its dual, an iff with its sides swapped --- valid for every $P$, so the
argument is made once per rule rather than once per row, and no Isabelle or Rocq
installation enters the construction. Neither form of warrant is a weakened
version of the other: a per-row obligation catches a rewriter that misfires on a
single statement, and a per-rule argument holds for every instance without
depending on a tactic finding the proof.

The semantic rules are labelled by their construction. \texttt{negate} replaces
the statement by its negation, which contradicts it for every $P$.
\texttt{forall\_to\_exists} yields a strictly weaker statement, which differs from
the original unless the domain is a singleton. \texttt{bump\_const} moves a
numeral and \texttt{strictness} loosens or tightens an inequality; both are
Lean-only, because they pick their target by position and the three
formalizations were written independently over different libraries, so the same
rule lands on a different subterm in each language. Over the problems Lean and
Isabelle share, $63\%$ differ in how many numerals their statement contains.
\begin{table}[H]
\centering
\caption{The PutnamBench slices. Controls are labelled \texttt{confirmed},
semantic mutations \texttt{disputed}. A dash marks a rule that is not applied in
that language; see the text.}
\label{tab:putnam-rules}
\small
\begin{tabular}{@{}llrrr@{}}
\toprule
& rule & Lean & Isabelle & Rocq \\
\midrule
\multirow{5}{*}{\textbf{control}}
& \texttt{true\_and}          &  98 &  94 &  62 \\
& \texttt{or\_false}          &  98 &  96 &  66 \\
& \texttt{double\_neg}        &  97 &  95 &  57 \\
& \texttt{forall\_not\_exists} &  24 &  22 &  38 \\
& \texttt{iff\_comm}          &  21 &  18 &  25 \\
\addlinespace
\multirow{4}{*}{\textbf{semantic}}
& \texttt{negate}             & 293 & 285 & 185 \\
& \texttt{bump\_const}        & 229 & --- & --- \\
& \texttt{strictness}         & 114 & --- & --- \\
& \texttt{forall\_to\_exists}  &  29 &  25 &  38 \\
\addlinespace
& untouched originals         & 293 & 285 & 185 \\
\midrule
& total                       & 1{,}296 & 920 & 656 \\
\bottomrule
\end{tabular}
\end{table}

\section{How \texttt{claimcheck} Works}
\label{sec:method}

\texttt{claimcheck} is a tool for checking whether a formal artifact means what
its natural-language requirement asks for. Given the pair, it returns a verdict,
\texttt{confirmed} or \texttt{disputed}, and names the discrepancy where it
finds one. Figure~\ref{fig:overview} and Figure~\ref{fig:example-dafny} show one
run each, on two Dafny contracts.

A run is two model calls. \emph{Informalize} receives the artifact and whatever
context is needed to read it, and writes an informalization: a plain-English
account of what the artifact guarantees. It is not given the requirement.
\emph{Compare} then receives two descriptions of the same thing in the same
language, the requirement and that informalization, and returns the verdict.
Both figures set the two calls out in order, with what each one reads.

Comparing the two English descriptions is the idea. One says what was asked
for, the other says what the artifact actually guarantees, and a difference in
meaning surfaces as a difference between two sentences rather than between a
sentence and a block of code. The contract in Figure~\ref{fig:overview} keeps
every word of its requirement and is satisfied by returning zero, and the
contract in Figure~\ref{fig:example-dafny} has exactly the right shape for its
requirement and one wrong index inside the slice. The informalization is what
makes that comparison possible, and it can only do the job because it was
written blind. An informalization produced with the requirement in view would
tend to echo it, and a near miss would read as a hit.

We also report a variant in which the judge is handed the artifact alongside the
informalization, to check the account against the source; it is the weaker of
the two on every language we test (Section~\ref{sec:results}).

\begin{figure}
\begin{trace}{\textbf{Dafny}\quad\texttt{remove\_front}, the slice takes the wrong end}
\stagehead{requirement}{}
Removes the front element from the array \texttt{a} and returns the resulting
array \texttt{c}.\\
\addlinespace[3pt]
\stagehead{contract}{}
\texttt{method remove\_front(a: array<int>) returns (c: array<int>)}\newline
\texttt{\ \ requires a.Length > 0}\newline
\texttt{\ \ ensures a[0..a.Length - 1] == c[..]}\\
\midrule
\stagehead{step 1 \ \ informalize}{reads the contract, never the requirement}
``The returned array \texttt{c} contains the elements of array \texttt{a} from
index 0 to \texttt{a.Length - 1} (i.e.\ all elements except the last one).''\\
\midrule
\stagehead{step 2 \ \ compare}{reads the requirement and the informalization}
\textbf{disputed}, \texttt{wrong-property}\newline
``The requirement says to remove the FRONT (first) element, but the
back-translation describes removing the LAST element (slice from index 0 to
\texttt{a.Length-1} excludes the last element, not the first).''\\
\end{trace}
\caption{An example \texttt{claimcheck} run on a Dafny contract. The contract
has the right shape and one wrong index inside the slice: \texttt{a[0..a.Length
- 1]} drops the last element where \texttt{a[1..]} would drop the first, so the
method it describes returns the array with the wrong end removed. The contract
verifies against a body that satisfies it.}
\label{fig:example-dafny}
\end{figure}

\section{Experimental Setup}
\label{sec:setup}

\paragraph{Baseline.} The baseline gets one row per call: the requirement, the
formal artifact and its context. It replies through a tool call whose only field
is \texttt{YES} or \texttt{NO}. The prompt contains no system message, worked
examples, reasoning instruction or failure vocabulary. Each of those is itself a reasoning aid, so
giving them to the baseline would make the comparison measure the prompt rather
than the tool. \emph{Baseline + list} is the one exception, and it is a control:
the same single call, given the judge's list of rewrites that leave a claim
unchanged and of the kinds of mismatch to look for. That block is taken verbatim
from the judge's own tool rather than written for this arm, so no wording in it
is ours to choose; the cost is that it mentions fields a bare
\texttt{YES}/\texttt{NO} tool does not have. It still answers \texttt{YES} or
\texttt{NO}, so what it holds fixed against \texttt{claimcheck} is the
information in the prompt. What separates the two-step pipeline from the output
schema it answers through is the no-back-translation ablation below, a single
call through the judge's own tool, fields and all, and so without that
mismatch.

\paragraph{\texttt{claimcheck}.} The two steps are those of
Section~\ref{sec:method}. The version we evaluate adds three things to the prompts the tool shipped with. The informalizer is given a short list of
structure that carries no meaning, such as a double negation or a conjunction
with an always-true clause, and describes the artifact with it discarded, so two
artifacts that say the same thing get the same informalization. The judge's tool
lists the rewrites that leave a claim unchanged in each language (in Dafny, for
example, two \texttt{ensures} clauses merged into one with \texttt{\&\&}, which
changes nothing because Dafny conjoins them by definition) and names the kinds
of mismatch to look for, such as a wrong constant or a missing condition. And
the judge is asked plainly whether the two express the same guarantee, where the
original wording told it to be strict and to prefer flagging a mismatch. The
variant we also report hands the judge the formal artifact alongside the
informalization; everything else, the tool schema included, is unchanged.

\paragraph{Settings.} Informalizer \texttt{claude-haiku-4.5}, judge
\texttt{claude-sonnet-4.6}, $8192$ max tokens, temperature $0$. Both return
their output through a tool call, and the judge's verdict is a single boolean. The rewrite list and the error vocabulary
are per language, following each slice's own construction: the judge answers
through a contract-shaped taxonomy on Dafny, whose categories name which clause
moved and in which direction, and a proposition-shaped one on Lean, Isabelle and
Rocq. Both steps get one row per call, as the baseline does, so the two see the
same input and any difference comes from the tool. Neither prompt carries the
row's name, which in earlier runs could reveal how a row was built
(Appendix~\ref{app:earlier-results}). Each run is scored on the rows it
answered; no run leaves more than 12 unanswered.

\paragraph{Ablations and sweep.} Each ablation changes one part of the tool.
\emph{Held-out rules} removes some rewrites from the lists, to test whether they
act as an answer key for the benchmark's controls: the always-true conjunction
and always-false disjunction on Lean, Isabelle and Rocq, and a shifted bound, a
swapped equality and a contrapositive on Dafny. \emph{No back-translation} skips
the informalize step, so the judge compares the requirement with the artifact
directly, under the same prompt otherwise. \emph{Informalizer sees requirement}
gives the informalizer the requirement it is otherwise denied. The first and
last are run on \texttt{claimcheck} as reported; no back-translation is run on
the variant that sees the artifact, since a judge with neither the artifact nor
an informalization would have nothing to read. The sweep repeats the baseline,
baseline + list and \texttt{claimcheck} with a different judge, the informalizer
fixed at \texttt{claude-haiku-4.5}: \texttt{gemini-3.7-flash},
\texttt{gpt-5.6-luna} and \texttt{deepseek-v4-pro}, reached through OpenRouter at
temperature $0$. \texttt{deepseek-v4-pro} runs with its reasoning mode off, since
it rejects a forced tool call otherwise. The last row of
Table~\ref{tab:sweep} (Appendix~\ref{app:sweep}) uses \texttt{gemini-3.7-flash}
in both roles.

\section{Results}
\label{sec:results}

\subsection{Main results}
\begin{table}[H]
\centering
\caption{Main results. Informalizer \texttt{claude-haiku-4.5}; judge and baseline \texttt{claude-sonnet-4.6}. Macro recall (mean of the \texttt{confirmed} and \texttt{disputed} recalls) $\pm$ its 95\% interval.}
\label{tab:results-main}
\small
\begin{tabular}{@{}lcccc@{}}
\toprule
 & Lean & Isabelle & Rocq & Dafny \\
\midrule
zero-shot baseline & 79.7 {\scriptsize$\pm$2.1} & 81.5 {\scriptsize$\pm$1.9} & 79.6 {\scriptsize$\pm$2.3} & 84.0 {\scriptsize$\pm$3.8}  \\
baseline + list & 83.2 {\scriptsize$\pm$1.9} & 85.3 {\scriptsize$\pm$1.8} & 83.0 {\scriptsize$\pm$2.3} & 88.9 {\scriptsize$\pm$3.2}  \\
claimcheck, judge sees artifact & 84.8 {\scriptsize$\pm$1.9} & 87.2 {\scriptsize$\pm$1.8} & 83.5 {\scriptsize$\pm$2.3} & 91.7 {\scriptsize$\pm$2.8} \\
claimcheck & 87.9 {\scriptsize$\pm$1.8} & 91.3 {\scriptsize$\pm$1.7} & 88.0 {\scriptsize$\pm$2.1} & 93.1 {\scriptsize$\pm$2.6} \\
\bottomrule
\end{tabular}
\end{table}

\texttt{claimcheck} is ahead of both baselines on all four languages: by $8.2$
to $9.8$ points of macro recall over the zero-shot baseline, and by $4.2$ to
$6.0$ over the same call handed the rewrite list
(Table~\ref{tab:results-main}). Most of the gain is on faithful rows. Confirmed
recall rises from $66.5$--$85.6\%$ with the list to $77.8$--$91.7\%$, on the
untouched originals as well as on the meaning-preserving controls: the list
baseline accepts $69.7$--$91.9\%$ of the originals and $64.1$--$82.2\%$ of the
controls, \texttt{claimcheck} $78.4$--$95.2\%$ and $77.4$--$89.8\%$. Disputed
recall moves less. It
rises on Dafny, from $92.3$ to $94.5\%$, falls by at most $3.3$ points on
Isabelle and Rocq, and falls on Lean, from $95.9$ to $88.4\%$.

Paired on the same rows, \texttt{claimcheck} fixes more of the list baseline's
errors than it introduces on every language: $148$ against $91$ on Lean (exact
McNemar $p = 3\times10^{-4}$), $128$ against $46$ on Isabelle
($p = 4\times10^{-10}$), $83$ against $37$ on Rocq ($p = 3\times10^{-5}$) and
$24$ against $9$ on Dafny ($p = 0.014$).

Handing the judge the artifact as well costs $1.4$ to $4.5$ points. That
variant disputes more of the faithful rows: its confirmed recall is $9.9$ to
$12.0$ points lower on Lean, Isabelle and Rocq, and $2.8$ lower on Dafny.

\subsection{Ablations}
\begin{table}[H]
\centering
\caption{Ablations, with the models of Table~\ref{tab:results-main}. Macro recall (mean of the \texttt{confirmed} and \texttt{disputed} recalls) $\pm$ its 95\% interval.}
\label{tab:ablations}
\small
\begin{tabular}{@{}lcccc@{}}
\toprule
 & Lean & Isabelle & Rocq & Dafny \\
\midrule
held-out rules & 88.0 {\scriptsize$\pm$1.8} & 91.8 {\scriptsize$\pm$1.7} & 87.5 {\scriptsize$\pm$2.3} & 92.3 {\scriptsize$\pm$2.8}  \\
no back-translation (judge sees artifact) & 82.0 {\scriptsize$\pm$2.0} & 81.9 {\scriptsize$\pm$2.0} & 77.5 {\scriptsize$\pm$2.6} & 86.7 {\scriptsize$\pm$3.4}  \\
informalizer sees requirement & 85.3 {\scriptsize$\pm$1.9} & 89.9 {\scriptsize$\pm$2.1} & 87.2 {\scriptsize$\pm$2.6} & 90.1 {\scriptsize$\pm$3.1} \\
\bottomrule
\end{tabular}
\end{table}

Without the back-translation, the judge compares the requirement with the
artifact directly and loses $2.8$ to $6.0$ points against the variant it is run
on (Table~\ref{tab:ablations}); on every language it ends below the single-call
baseline with the list. Giving the informalizer the requirement costs $0.8$ to
$3.0$ points, and the loss is on the disputed side: disputed recall falls by
$6.0$ to $8.8$ points while confirmed recall holds or rises. That is the
direction Section~\ref{sec:method} predicts for an informalization written with
the requirement in view. Holding some of the rules that generate the
benchmark's controls out of the lists moves macro recall by at most $0.8$
points, inside every interval, so the lists are not acting as an answer key for
those rules.

\subsection{Model sweep}

% fig-sweep.pdf is generated by fig-sweep.py from tables/sweep.tex and
% tables/main.tex; rerun it whenever either table changes.
\begin{figure}[H]
\centering
\includegraphics[width=\linewidth]{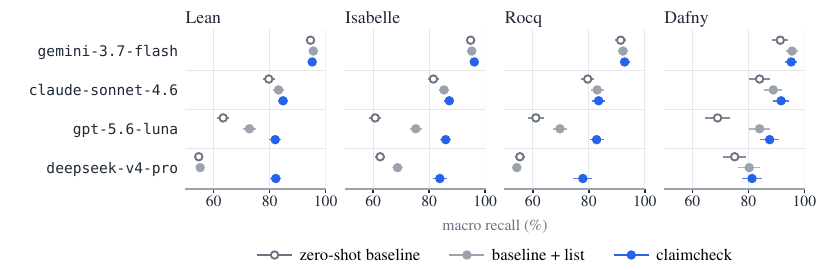}
\caption{The model sweep, with $95\%$ intervals; values in
Table~\ref{tab:sweep}. As there, \texttt{claimcheck} is the variant whose judge
also sees the artifact, and the \texttt{claude-sonnet-4.6} row is
Table~\ref{tab:results-main}'s judge under that variant.}
\label{fig:sweep}
\end{figure}

The sweep's \texttt{claimcheck} rows are the variant whose judge also sees the
artifact, and how much they gain depends on the judge (Figure~\ref{fig:sweep};
the values are in Table~\ref{tab:sweep}, Appendix~\ref{app:sweep}).
For the two weaker judges the gain is large: over the zero-shot baseline,
\texttt{gpt-5.6-luna} gains $18.6$ to $25.2$ points and
\texttt{deepseek-v4-pro} $6.2$ to $27.5$, while the list alone adds at most
$15.0$. \texttt{gemini-3.7-flash} already reaches $91.4$ to $94.8$ in a single
call; it gains $0.6$ to $3.9$ over that, and is within a point of its list
baseline on every language. Using it as the informalizer too changes macro
recall by at most $1.1$ points. \texttt{claimcheck} narrows the gap between
judges: the spread across the three falls from $22.4$--$39.9$ points for the
baseline to $12.3$--$14.9$.

\section{Related Work}
\label{sec:related}

\paragraph{Specifications that verify but say the wrong thing.}
Boehm's distinction between verification, building the product right, and
validation, building the right product, is forty years old
\citep{boehm1984vv}, and a verifier does only the first. Tools that write
specifications with a model mostly take the verifier's word: SpecGen
\citep{ma2024specgen} and AutoSpec \citep{wen2024autospec} keep whichever
candidate contract the verifier accepts, which says nothing about whether the
accepted candidate is the one the author meant. Where the specification is
scored at all, it is scored against code. nl2postcond asks whether the
reference implementation passes a generated postcondition and how many buggy
versions of that implementation the postcondition rejects
\citep{endres2024nl2postcond}, and Richter and Wehrheim do the same
for full contracts \citep{richter2025contracts}. Benchmarks for verified code
generation have met the gap from the other side. The authors of the vericoding
benchmark read their specifications by hand and found about a tenth too weak and
a further sixth poorly translated from the task \citep{bursuc2025vericoding}.
AlphaVerus adds an exploit step that tries to write a trivial program satisfying
the specification, because a model rewarded for verified code will otherwise
learn to weaken the specification \citep{aggarwal2025alphaverus}. Vacuity
detection in model checking is the older form of the same worry: a temporal
property can hold of a system for a reason that has nothing to do with what it
was written to say \citep{beer2001vacuity, kupferman2003vacuity}. All of these
check the specification against code or traces. \texttt{claimcheck} checks it
against the requirement.

\paragraph{Checking against a reference formalization.}
When a trusted formalization exists, faithfulness becomes equivalence between
two formal objects, which can be decided or at least attempted. CLEVER asks the
model to prove its specification equivalent to a hidden one
\citep{thakur2025clever}; Verina scores soundness and completeness against a
ground-truth specification, by proof where it goes through and by tests where it
does not \citep{ye2025verina}. In mathematics, Poiroux et al.\ check
derivability in both directions in Lean \citep{poiroux2024reliable}, LeanEuclid
hands the equivalence to an SMT solver \citep{murphy2024leaneuclid}, and GTED
compares operator trees \citep{liu2025gted}. These are precise where they apply,
and they apply only once someone has written the answer down. Our rows carry no
reference beyond the requirement. The controls of Section~\ref{sec:benchmark} do
the one job that can be done without one, which is to certify that a rewrite of
the artifact preserved its meaning: by proof where a prover decides it, and by
an argument about the rule where none runs.

\paragraph{Judging against the requirement directly.}
Without a reference the judge has to read the natural language, and the usual
design shows a model both sides and asks. ProofFlow scores faithfulness this way,
clause by clause \citep{cabral2025proofflow}; Zhang et al.\ take the agreement
of two such judges as a stand-in for human review of Lean statements
\citep{zhang2026beyondcompilation}. The record of the direct judge is mixed. A GPT-4 judge shown an informal
statement and its formalization had far lower precision than FormalAlign's
trained scorer \citep{lu2024formalalign}, and Verus-SpecGym finds that a judge misses a quarter
of the specification failures its executable evaluator catches
\citep{agarwal2026specgym}. The literature on model judges gives reasons to
expect this. What a judge sees next to the thing it is grading shapes the
verdict: candidates are favoured based on position \citep{wang2024notfair}, as is the
judge's own output \citep{panickssery2024self}; a prior score shown as metadata
anchors the grade even when it is uninformative by construction
\citep{kapetanovic2026anchoring}; and a supplied reference answer is weighed
against what the judge already believes rather than taken as given
\citep{lee2026reference}. Our baseline is this direct judge with nothing else in the prompt, and the
requirement is the anchor \texttt{claimcheck} withholds from the informalizer.

\paragraph{Informalization and round trips.}
The step \texttt{claimcheck} rests on is not new. Wu et al.\ noticed that Codex
translated Isabelle into English far more reliably than the other way and
proposed back-translation as a source of training data
\citep{wu2022autoformalization}, which MMA then produced at scale
\citep{jiang2024multilanguage}. Lean Workbook uses it as a filter: each formal
statement is translated back, and a model is asked whether the two English texts
pose the same problem \citep{ying2024leanworkbook}. Li et al.\ replace the model
with embedding similarity between the original and its back-translation
\citep{li2024symbolic}. Two recent systems are close. Amrollahi et al.\
formalize a statute, back-translate it, formalize it again, and let an SMT
solver decide whether the two formalizations agree, calling a model only to
diagnose which step failed \citep{amrollahi2026roundtrip}. Monty, for JML
assertions, has one model describe the generated assertion and another compare
that description with the original clause by clause \citep{liu2026faithful}.
Outside program verification, nl2spec and ARTEMIS put the round trip in front of
the user for temporal logic, showing sub-translations or distinguishing traces
to accept or reject \citep{cosler2023nl2spec, mendoza2026artemis}. Each of these
uses the round trip inside its own system and reports the system's end-to-end
gain. Clover is the closest, and it is where our Dafny slice comes from: it
regenerates a docstring from the annotations without seeing the original and
asks GPT-4 whether the two match \citep{sun2024clover}. It scores that check
alone, accepting all 60 originals it evaluates and rejecting 40 to 70\% of each incorrect
class with a single attempt, but never against a model shown both sides; and on its rows, where every
faithful row is untouched, detecting edits would score the whole faithful half.
Mathesis makes that comparison, with DeepSeek-V3 throughout, on model-written
Lean statements labelled by hand, and the round trip lost on F1, $0.65$ to
$0.85$ \citep{yu2025mathesis}. Our controls remove the edit signal, and
Section~\ref{sec:results} runs the comparison across four languages and four
judges, against a single call given the same information.
\paragraph{Benchmarks with labelled mismatches.}
Labelled pairs of requirement and formalization are scarce, and the ones that
exist are made by damage. CloverBench mutates the docstring, the annotations or
the code \citep{sun2024clover}; ProofNetVerif labels model-written
formalizations against a reference \citep{poiroux2024reliable}; DriftBench
plants controlled drift in Lean statements \citep{mohammad2026driftbench}.
FaithformBench scores the unperturbed rows next to the perturbed ones
\citep{cornish2026faithform}, but its unperturbed rows are untouched, and ours
are rewritten. The construction descends from mutation testing, which has a
known hazard: the equivalent mutant, an edit that changes the text and not the
behaviour, which must be found and discarded or it corrupts the score
\citep{jia2011mutation}. We keep ours and label them faithful. Certifying one is
an equivalence question, and our four slices answer it differently: the Dafny
verifier decides it per row, a fixed tactic list decides it per row in Lean, and
in Isabelle and Rocq the rules kept are propositional schemas whose correctness
is argued once for the rule rather than once for the row.

\section{Limitations}
\label{sec:limitations}

Our benchmark may not be hard enough to separate systems clearly: a single call
already reaches between $79.6$ and $84.0$ macro recall, and between $83.0$ and
$88.9$ once it is handed the rewrite list, so the room a two-step tool can occupy
is small and differences between systems compress into a few points. Its
composition narrows that room further. In Isabelle and Rocq \texttt{negate} is
$92\%$ and $83\%$ of the disputed rows, and asserting the negation of a statement
is the easiest defect in the benchmark to catch, so what separates systems on
those two slices is almost entirely the confirmed side; and there the three
trivial wrappers are about $45\%$ of the rows, which tests whether a judge is
disturbed by structure carrying no meaning rather than whether it can read a
specification. Lean and Dafny carry defects of several strengths and are where
the disputed half discriminates.

The labels also carry different warrants. Every control was shown to preserve
meaning, by the Dafny verifier, by a tactic list in Lean, or by a propositional
argument in Isabelle and Rocq, but the semantic mutations rest on the rule that
produced them, and for \texttt{bump\_const} and \texttt{strictness} that is
intent alone, since a moved numeral or a loosened bound will in rare cases leave
a statement equivalent to the one it replaced. And all four slices are single
contracts or single theorem statements, short enough that a requirement and its
formalization can be held in view at once. Artifacts whose meaning lives in
their structure rather than in a proposition, and developments too long to read
in one pass, are outside what we measure; finding such tasks is the main thing we
would change before making a stronger claim.

\section{Conclusion}
\texttt{claimcheck} checks whether a formal artifact means what the
natural-language requirement it was written from asks for, which is the
question a verifier never asks. It does so in two model calls: one
informalizes the artifact without seeing the requirement, the other compares
the two English texts. Across Dafny contracts and Lean, Isabelle and Rocq
theorem statements it gains $8.2$ to $9.8$ points of macro recall over a direct
judge, and $4.2$ to $6.0$ over the same judge handed the same list of
meaning-preserving rewrites in a single call. Undoing any of its three choices lowers the score: removing the back-translation costs $2.8$ to $6.0$ points,
letting the informalizer see the requirement up to $3.0$, and showing the judge
the artifact up to $4.5$.

\FloatBarrier
\section*{Acknowledgments}
Claimcheck was initially developed as part of a Claude Code hackathon; we thank Anthropic for their support in the context of that. We also thank Apart Research and their Secure Program Synthesis fellowship organizers, especially Eitán Sprejer,  Max von Hippel and Quinn Dougherty. Simon Henniger was supported by the Harvard John A. Paulson School of Engineering and Applied Sciences (SEAS) Prize Fellowship and the German Academic Fellowship Organization, funded by the German Federal Ministry for Economic Affairs and Energy.

\section*{Code Availability}

Our code is available at \url{https://github.com/metareflection/claimcheck}.

\bibliographystyle{plainnat}
\bibliography{refs}
%%%%%%%%%%%%%%%%%%%%%%%%%%%%%%%%%%%%%%%%%%%%%%%%%%%%%%%%%%%%

\appendix

\clearpage
\section{Model sweep table}
\label{app:sweep}

The values plotted in Figure~\ref{fig:sweep}, with the \texttt{gemini-3.7-flash}
pairing in both roles as an extra row.

\begin{table}[H]
\centering
\caption{Model sweep: other judges, with informalizer \texttt{claude-haiku-4.5} unless stated. Macro recall (mean of the \texttt{confirmed} and \texttt{disputed} recalls) $\pm$ its 95\% interval.
The \texttt{claimcheck} rows are the variant whose judge also sees the artifact.}
\label{tab:sweep}
\small
\begin{tabular}{@{}lcccc@{}}
\toprule
 & Lean & Isabelle & Rocq & Dafny \\
\midrule
\texttt{gemini}: baseline & 94.6 {\scriptsize$\pm$1.2} & 94.8 {\scriptsize$\pm$1.2} & 91.4 {\scriptsize$\pm$1.9} & 91.4 {\scriptsize$\pm$2.9}  \\
\texttt{gemini}: baseline + list & 95.6 {\scriptsize$\pm$1.1} & 95.2 {\scriptsize$\pm$1.2} & 92.2 {\scriptsize$\pm$1.8} & 95.6 {\scriptsize$\pm$2.1}  \\
\texttt{gemini}: claimcheck & 95.2 {\scriptsize$\pm$1.1} & 96.1 {\scriptsize$\pm$1.1} & 92.8 {\scriptsize$\pm$1.8} & 95.3 {\scriptsize$\pm$2.2} \\
\midrule
\texttt{luna}: baseline & 63.4 {\scriptsize$\pm$2.1} & 60.7 {\scriptsize$\pm$2.1} & 61.1 {\scriptsize$\pm$2.8} & 69.0 {\scriptsize$\pm$4.6} \\
\texttt{luna}: baseline + list & 72.8 {\scriptsize$\pm$2.2} & 75.2 {\scriptsize$\pm$2.2} & 69.7 {\scriptsize$\pm$2.6} & 84.0 {\scriptsize$\pm$3.8}  \\
\texttt{luna}: claimcheck & 82.0 {\scriptsize$\pm$2.1} & 85.9 {\scriptsize$\pm$2.0} & 82.8 {\scriptsize$\pm$2.5} & 87.6 {\scriptsize$\pm$3.3}  \\
\midrule
\texttt{deepseek}: baseline & 54.7 {\scriptsize$\pm$1.2} & 62.5 {\scriptsize$\pm$1.8} & 55.4 {\scriptsize$\pm$1.7} & 75.1 {\scriptsize$\pm$4.2}  \\
\texttt{deepseek}: baseline + list & 55.2 {\scriptsize$\pm$1.2} & 68.7 {\scriptsize$\pm$2.0} & 54.3 {\scriptsize$\pm$1.3} & 80.3 {\scriptsize$\pm$3.9}  \\
\texttt{deepseek}: claimcheck & 82.2 {\scriptsize$\pm$1.9} & 83.8 {\scriptsize$\pm$2.6} & 77.9 {\scriptsize$\pm$3.4} & 81.3 {\scriptsize$\pm$3.5}  \\
\midrule
\texttt{gemini} in both roles & 94.7 {\scriptsize$\pm$1.2} & 95.0 {\scriptsize$\pm$1.3} & 91.7 {\scriptsize$\pm$1.9} & 95.3 {\scriptsize$\pm$2.2}  \\
\bottomrule
\end{tabular}
\end{table}

\section{Variant examples}
\label{app:variants}

\begin{table}[H]
\centering
\caption{Every Dafny variant on one Clover program, \texttt{swap}. \texttt{C1}
is the only class where the requirement is what is wrong: the contract is still
a correct swap, but the requirement now asks for something else. \texttt{C2}
narrows the contract to inputs where \texttt{X==0}. \texttt{C6} returns the
inputs unchanged. The control merges two \texttt{ensures} clauses into one,
which changes the text but not the meaning.}
\label{tab:variant-examples}
\small
\begin{tabular}{@{}p{2.8cm}p{3.4cm}p{4.6cm}l@{}}
\toprule
variant & requirement & contract & label \\
\midrule
original & Swap two integers. & \texttt{ensures x==Y}\newline\texttt{ensures y==X} & \texttt{confirmed} \\
docstring mutated (\texttt{C1}) & Assign the second integer to the first integer. & unchanged & \texttt{disputed} \\
contract mutated (\texttt{C2}) & Swap two integers. & \texttt{requires X==0}\newline\texttt{ensures x==Y}\newline\texttt{ensures y==X}\newline\texttt{ensures y==0} & \texttt{disputed} \\
contract + body (\texttt{C6}) & Swap two integers. & \texttt{ensures x==X}\newline\texttt{ensures y==Y} & \texttt{disputed} \\
control, meaning-preserving (\texttt{merge\_ensures}) & Swap two integers. & \texttt{ensures (x==Y) \&\& (y==X)} & \texttt{confirmed} \\
\bottomrule
\end{tabular}
\end{table}

\paragraph{Why \texttt{C1} is kept and \texttt{C3} is not.} \texttt{C1} has a
polarity the rest of the benchmark lacks: the docstring is the thing that is
wrong, so the pair disagrees because of its requirement. The fourth released
class, \texttt{C3}, weakens docstring and contract in tandem. The pair then
often still agrees with itself and would be \texttt{confirmed} for our question
even though Clover rightly calls it incorrect for its own, and that agreement is
not reliable enough to label the class by rule in either direction. Its 60 rows
are therefore left out of the benchmark, and labelling them by hand is future
work. Clover's own annotation of \emph{how} each variant diverges is retained on
the rows we keep, so a directional question such as whether a contract
guaranteeing more than its docstring claims should count as unfaithful can be
asked as a filter rather than baked into the labels.

\section{Earlier results (before the row-name fix)}
\label{app:earlier-results}

Row identifiers in our data record how each row was built:
\texttt{putnam\_1962\_a1\_\_control\_\_true\_and} on the PutnamBench slices,
\texttt{abs\_\_C1} on Dafny. An earlier version of the harness rendered that
identifier as the artifact's name inside both \texttt{claimcheck} prompts, while
the baseline prompt never received it, so one arm could read the label off a
string the other never saw. The direction of such a leak is not obvious --- a
judge that trusts a \texttt{control} tag accepts rows it should accept, and a
\texttt{semantic} tag does the reverse --- so it does not simply inflate a
score, it makes it uninterpretable.

Every figure in this paper was produced after that identifier was cut back to
its base form, and no row's name, artifact or context now records how the row
was made. The earlier runs are superseded and we do not reproduce them here.
They differed in three further ways: the judge was handed the formal artifact
alongside the informalization and told to be strict, which is the variant
Section~\ref{sec:results} now reports second; rows were batched rather than sent
one per call; and the evaluation covered a SQL slice derived from Spider, which
the benchmark of Section~\ref{sec:benchmark} no longer contains. The released
code and data include that slice and those runs.

\end{document}